\documentclass[conference]{IEEEtran}

\usepackage{cite}
\usepackage{amsmath,amssymb,amsfonts}
\usepackage{graphicx}
\usepackage{booktabs}
\usepackage{textcomp}
\usepackage{xcolor}
\def\BibTeX{{\rm B\kern-.05em{\sc i\kern-.025em b}\kern-.08em
    T\kern-.1667em\lower.7ex\hbox{E}\kern-.125emX}}

\begin{document}

\title{Vision Guided Target Conditioned Control for Autonomous Excavation}

\author{
\IEEEauthorblockN{
Shuai Zhao$^{1}$,
Ji-An Pan$^{2,*}$,
Junwei Li$^{2}$,
Xun Tang$^{3}$,
Fansen Xi$^{3}$,
Qing Xu$^{3}$,
Keqiang Li$^{3}$,
and Jianqiang Wang$^{3}$
}
\IEEEauthorblockA{
$^{1}$Liaoning University, Shenyang, China\\
$^{2}$Northeastern University, Shenyang, China\\
$^{3}$Tsinghua University, Beijing, China\\
$^{*}$Corresponding author: Ji-An Pan, panjian@me.neu.edu.cn
}
}

\maketitle

\begin{abstract}
Autonomous excavation requires an intelligent control system that can convert spatial work intent into coordinated bucket motion under contact-rich soil interaction.
This paper presents a target-conditioned intelligent control framework for autonomous excavation in a physics-based deformable-soil simulation workflow.
An image-aligned target mask serves as a visual spatial command for the desired digging region, while a mask-conditioned Action Chunking Transformer maps multi-view RGB observations, proprioception, and the target mask to temporally extended joystick commands.
To reduce target-ignoring behavior, demonstrations are organized with paired-condition supervision, where the same or closely matched scene is demonstrated with different target masks and corresponding action chunks.
The framework is evaluated through both a diagnostic manipulation task and an excavation simulation benchmark with single-scoop and sequential pile-clearing protocols.
In manipulation, target success is 4\% for no-condition ACT, 63\% for non-paired mask-conditioned ACT, and 96\% for paired-condition mask-conditioned ACT.
In sequential pile clearing, paired-condition mask-conditioned ACT removes 76.8\% of the pile versus 27.4\% and 15.7\% for the two baselines, with 91.0\% human-normalized efficiency.
The results show that visual target conditioning, paired demonstration structure, and action-chunk control form a practical cyber-physical simulation pipeline for excavator automation.
\end{abstract}


\section{Introduction}

Excavation automation can reduce operator exposure to hazardous worksites and alleviate shortages of skilled operators~\cite{huh2023trajectory}.
However, reliable control remains difficult because a fixed-base excavator must coordinate four work-control joints through a long-horizon dig-load-dump cycle under nonlinear and spatially varying soil interaction~\cite{huh2023trajectory,egli2022soil}.
Autonomy must therefore map a desired work region to a mechanically effective command sequence.

Imitation learning can acquire coordinated visuomotor behavior from demonstrations, but performance depends on both policy architecture and the demonstration distribution~\cite{mandlekar2021robomimic,florence2021implicit}.
Target-conditioned excavation adds a specific challenge: a policy may learn an effective default scoop yet remain unable to allocate successive scoops across a pile.
Conditional imitation learning shows that high-level commands can select behaviors unavailable from an unconditioned policy~\cite{codevilla2018conditional}.
ACT represents coordinated motion as temporally extended action sequences~\cite{zhao2023act}, but action chunking alone does not ensure that the policy follows a spatial target rather than scene-correlated behavior.

This paper presents a target-conditioned intelligent control framework for autonomous excavation in physics-based simulation.
An image-aligned mask specifies the desired dig region, while a mask-conditioned ACT policy maps multi-view RGB observations, proprioception, and the mask to joystick-command chunks.
Paired-condition supervision uses identical or closely matched scenes with different targets and corresponding action chunks to create direct target-action contrasts.
Evaluation proceeds from diagnostic manipulation to single-scoop and sequential pile-clearing protocols.
In pile clearing, the paired-condition policy removes 76.8\% of the pile, compared with 27.4\% for non-paired conditioning and 15.7\% without target conditioning.

\begin{figure*}[t]
\centering
\includegraphics[width=0.97\textwidth]{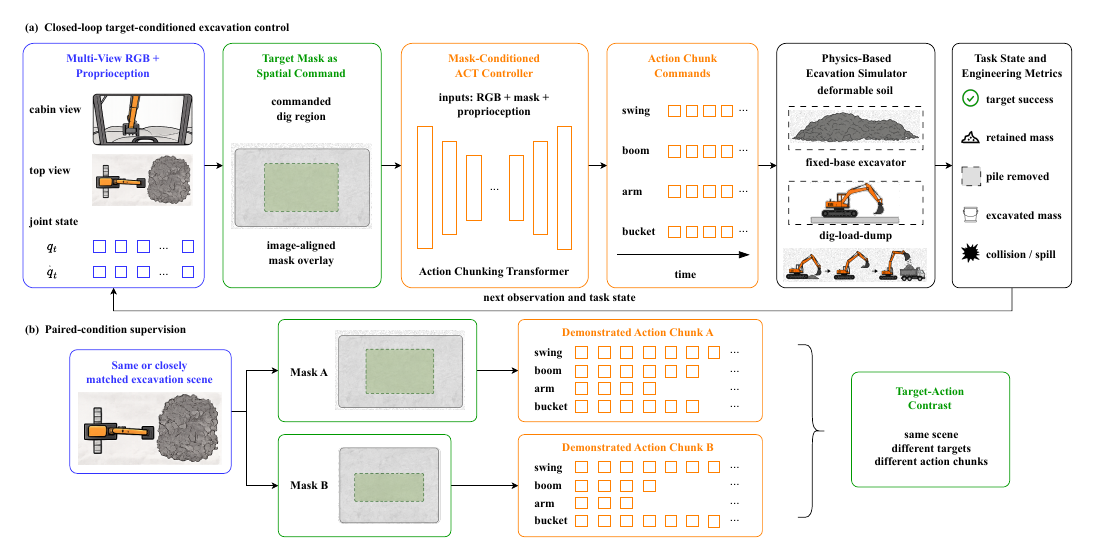}
\caption{Closed-loop target-conditioned intelligent control framework for autonomous excavation. (a) A visual target mask specifies the commanded dig region, and the mask-conditioned ACT controller maps multi-view observations and proprioception to action-chunk work-control commands in a physics-based excavation simulator. (b) Paired-condition supervision creates target-action contrasts by pairing the same or closely matched excavation scene with different target masks and corresponding demonstrated action chunks.}
\label{fig:framework_paired}
\end{figure*}

The paper makes four contributions.
First, it connects image-aligned spatial commands, multi-view sensing, proprioception, and four-dimensional action-chunk control in a deformable-soil simulator.
Second, it introduces paired-condition supervision to expose how the commanded region changes the required control sequence.
Third, it separates target compliance, single-scoop effectiveness, and sequential pile-clearing performance through material-removal and failure metrics.
Fourth, it compares no-condition, non-paired, and paired-condition policies and checks interface-level execution on a scaled physical testbed.
Fig.~\ref{fig:framework_paired} summarizes the resulting closed-loop architecture and the paired demonstration design used to connect visual spatial commands with executable excavation-control chunks.

\section{Related Work}

\subsection{Imitation Learning for Continuous Control}

Behavior cloning from demonstrations is widely used for visuomotor robot control, but its performance depends on both the policy architecture and the structure of the demonstration distribution~\cite{mandlekar2021robomimic,florence2021implicit}.
Sequence-level policies reduce compounding error by predicting temporally extended actions.
ACT predicts action chunks with a transformer policy~\cite{zhao2023act}, while Diffusion Policy generates action sequences through conditional denoising and receding-horizon execution~\cite{chi2024diffusion}.
This work uses ACT as the low-level controller and focuses on how target-conditioned demonstrations should be organized for excavator automation.

\subsection{Target-Conditioned Robot Learning}

Robot policies can be conditioned on visual goals, spatial prompts, language, or multimodal task descriptions~\cite{zeng2021transporter,jang2022bcz,jiang2023vima,brohan2023rt1}.
Such inputs can specify what the robot should do, but the policy must still learn how the condition modulates the necessary continuous action sequence.
In this paper, the target mask is not treated as a new prompt modality.
It is an engineering interface for spatial excavation intent and a controlled way to test whether the policy behavior changes with the commanded region.

\subsection{Autonomous Excavation and Intelligent Control}

Excavation is challenging because control is long-horizon, contact-rich, and coupled to deformable terrain.
Prior work has studied force estimation for excavators~\cite{SHEN2025111902}, autonomous excavation planning and execution systems~\cite{CHO2026106742}, reinforcement learning for excavation control~\cite{kurinov2020rl,osa2022depthrl}, and bucket-trajectory planning from terrain observations~\cite{huh2023trajectory}.
More broadly, recent intelligent control studies have examined deep learning and reinforcement learning for autonomous vehicles, finite-time state estimation under biased sensing, and distributed decision making for multi-agent systems~\cite{wen2025airship,wen2025attitude,Wen2025AnOO}.
These studies motivate excavation as an intelligent-control and robotics automation domain.
Our focus is a simulation-based target-conditioned control interface that maps a visual commanded region to continuous joystick-style work-control chunks.

\section{Closed-Loop Control Architecture}

\subsection{Control Problem and System Loop}

We consider fixed-base target-region excavation as an unmanned-system control problem.
At time $t$, the system receives observation $o_t$ and target condition $c$.
The observation includes multi-view RGB images and proprioceptive state.
The target condition is represented as an image-aligned binary mask over the main task view.
The closed-loop system alternates between sensing, target-conditioned action prediction, command execution, and task-state evaluation.
The policy predicts a chunk of future commands
\begin{equation}
    A_t = (a_t, a_{t+1}, \ldots, a_{t+K-1}) = \pi_\theta(o_t, c),
\end{equation}
where $K$ is the chunk length.
For excavation, each action $a_t$ is a four-dimensional joystick-style command for swing, boom, arm, and bucket.
From an intelligent-control perspective, the target mask acts as a spatial reference signal, while the learned action chunk acts as a short-horizon control command sequence.

\subsection{Target-Conditioned Action-Chunk Controller}

The target mask is concatenated with the aligned RGB image as a fourth channel.
In multi-view setups, the main task view receives the valid target mask, while auxiliary views are padded with zero-mask channels.
The mask-conditioned ACT policy encodes RGB-mask visual tokens and proprioceptive tokens, then decodes a temporally extended command chunk.
Action chunking is well suited to excavation, as useful behavior unfolds over coherent phases (e.g., approach, digging, curling, lifting, and dumping).

For a demonstrated trajectory with target condition $c$, observation $o_t$, and demonstrated future action chunk $A_t^*$, the policy is trained by supervised imitation:
\begin{equation}
    \mathcal{L}(\theta) =
    \sum_t \left\| \pi_\theta(o_t, c) - A_t^* \right\|_1
    + \lambda_{\mathrm{KL}} \mathcal{L}_{\mathrm{KL}},
\end{equation}
where $\lambda_{\mathrm{KL}}=10$ in the implementation.

\subsection{Paired Demonstration Design}

Standard target conditioning may prove insufficient if each scene is associated with only one target.
In such cases, the policy might memorize target-correlated features without learning a true causal dependence on the target condition.
Paired-condition supervision changes the data distribution.
For each scene group $i$, demonstrations are collected as
\begin{equation}
    \mathcal{D}_i = \{(o^{ij}_t, c_{ij}, A^{ij*}_t)\}_{j,t},
\end{equation}
where the same or closely matched initial context appears with multiple target masks $c_{ij}$ and corresponding action chunks $A^{ij*}_t$.
The within-scene contrast discourages scene-identity shortcuts because changing the mask changes the correct control sequence.
This design is important for excavation because a visually similar pile can support several locally valid scoops, and the desired scoop should be selected by the commanded work region rather than by scene appearance alone.

\section{Simulation Environment and Data Workflow}

The excavation experiments use a physics-based deformable-soil simulation workflow implemented around a fixed-base excavator with four work-control commands.
The deformable terrain is modeled as cohesionless sand with a density of $1474\,\mathrm{kg/m^3}$ and a friction angle of $39.0^\circ$.
Dynamics are advanced with a $0.02\,\mathrm{s}$ time step, and all compared methods share identical terrain and contact settings.
On top of the physics simulator, we implement controlled pile-state initialization, target-region overlays, joystick-command interfaces, and material-removal measurements.
The resulting testbed integrates teleoperation recording, replay-based quality control, offline ACT training, and live rollout evaluation.
The results therefore characterize policy differences under one nominal soil condition rather than calibrated prediction across field soil types.

Excavation demonstrations are collected by joystick operation.
For target-conditioned demonstrations, the operator executes a dig-load-dump cycle for the commanded region indicated by the target mask.
For non-paired data, each scene is associated with one commanded region.
For paired-condition data, the same or closely matched pile state is demonstrated with different commanded regions.
This provides a form of counterfactual supervision: the visual context remains largely identical, whereas the target mask and the correct joystick sequence vary.

Two excavation protocols are used.
Single-scoop trials evaluate whether the policy can execute one mechanically effective dig-load-dump primitive.
Sequential pile-clearing trials repeatedly assign local dig regions over the remaining pile and evaluate whether target-conditioned scoops can be composed to cover a larger pile.
The primary metrics are target success, retained deposited mass, pile removed percentage, and human-normalized efficiency relative to joystick demonstrations under the matched protocol.

\subsection{Implementation Workflow}

The simulation workflow is organized around a step-wise closed-loop interface between the controller and the physics-based simulator.
At each control step, the environment returns visual observations, proprioceptive state, and task-state variables.
The policy consumes the current observation and target mask, predicts an action chunk, and the runner executes the corresponding joystick-style commands through the simulator interface.
This design forms a cyber-physical simulation loop and allows the same policy abstraction to be used for teleoperation replay, offline training, and closed-loop evaluation.

Demonstration episodes are recorded with synchronized observations, action labels, timestamps, and task-state measurements.
The data workflow supports teleoperation recording, replay-based quality control, dataset inspection, ACT training, and rollout logging.
For excavation, the task state includes material-related variables such as bucket mass, excavated mass, deposited mass in the target box, distance-to-target quantities, hard target contact counts, and bucket depth relative to the dig area.
These variables are not direct policy outputs; they are used for evaluation, quality control, and success detection.

Target masks are defined in the main RGB coordinate frame and undergo the same cropping and resizing as the RGB input; RGB and depth are internally registered by the RGB-D camera.
For the manipulation dataset, Grounding DINO detections prompt SAM 2 to generate object masks.
For excavation, Qwen2.5-VL-7B combines the registered depth observation with geometric feasibility rules to select a local dig region, from which SAM 2 produces the aligned mask.
Masks are generated independently in each platform's image frame rather than transferred from simulation to the scaled testbed.
The no-condition baselines follow the same target schedule, but the mask is withheld from the policy input, isolating the effect of target conditioning.

\subsection{Engineering Metrics}

\begin{figure}[t]
\centering
\includegraphics[width=\linewidth]{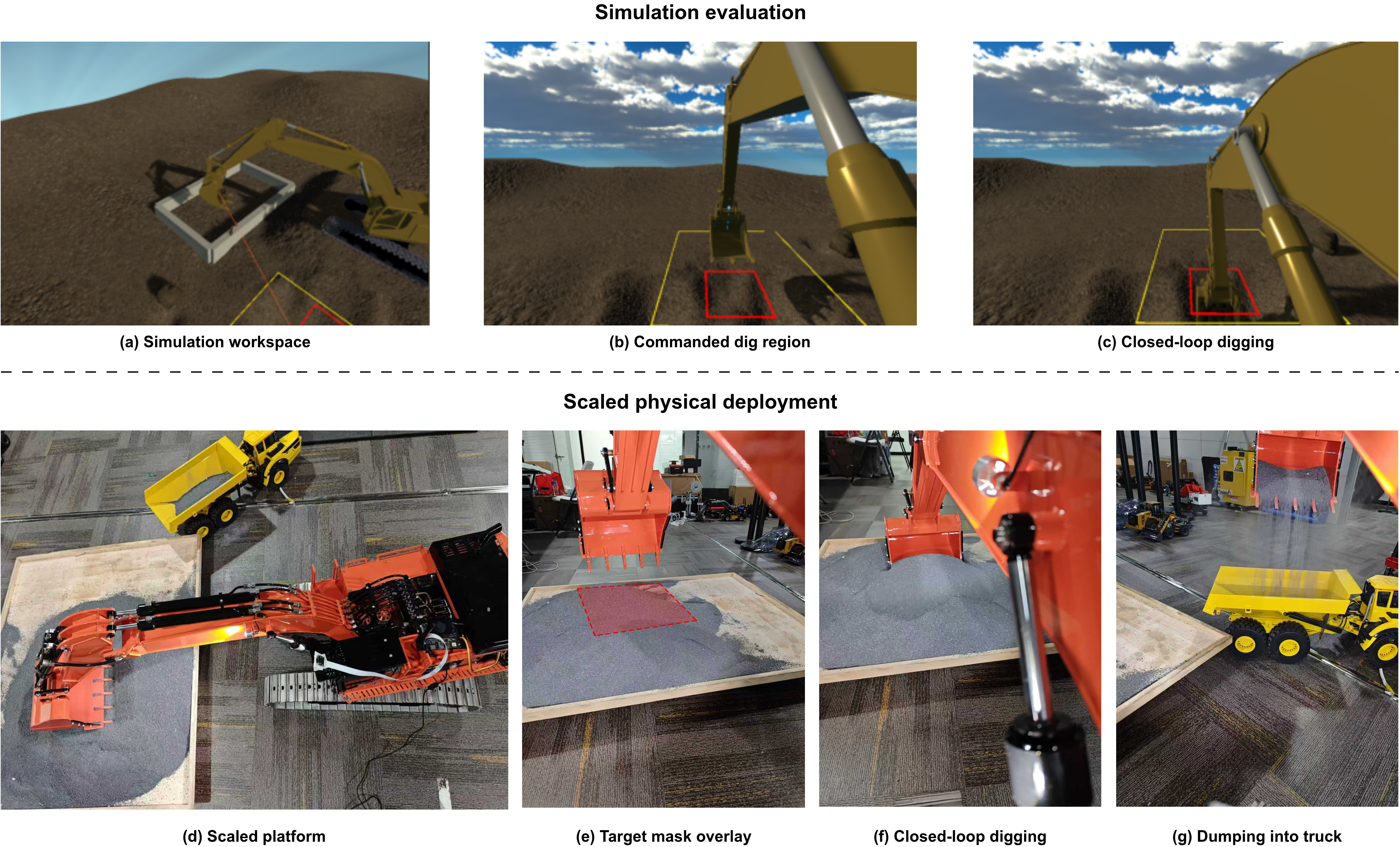}
\caption{Target-conditioned excavation evaluation. The physics-based simulation provides controlled deformable-soil evaluation with commanded dig regions and material-removal metrics. The scaled physical testbed is included only as an interface-level feasibility check; all quantitative excavation comparisons are reported in simulation.}
\label{fig:excavation_eval}
\end{figure}

The evaluation protocol is designed to capture both target compliance and excavation utility.
Single-scoop trials measure whether a policy can complete one local dig-load-dump cycle, while sequential pile-clearing trials measure whether repeated local commands can decompose a larger pile into several useful scoops.
Table~\ref{tab:metrics} summarizes the main task variables and metrics used by the simulation workflow, and Fig.~\ref{fig:excavation_eval} shows the corresponding excavation evaluation setup.

\begin{table}[t]
\caption{Engineering metrics for target-conditioned excavation evaluation.}
\label{tab:metrics}
\centering
\small
\resizebox{\linewidth}{!}{%
\begin{tabular}{ll}
\toprule
Metric or variable & Role in evaluation \\
\midrule
Target success & Loading begins in the commanded dig region \\
Retained mass & Deposited mass retained after a scoop cycle \\
Pile removed & Fraction of initial pile removed after matched cycles \\
Human-normalized efficiency & Performance relative to joystick demonstrations \\
Residual bucket mass & Checks whether dumping is complete \\
Hard target collision count & Strict safety and failure indicator \\
Spill-before-target event & Failure mode for premature material loss \\
Distance to target & State variable for approach and deposit analysis \\
Bucket depth below dig plane & State variable for dig initiation analysis \\
\bottomrule
\end{tabular}
}
\end{table}

\section{Experimental Validation}

\subsection{Diagnostic Manipulation Study}

\begin{figure}[t]
\centering
\includegraphics[width=\linewidth]{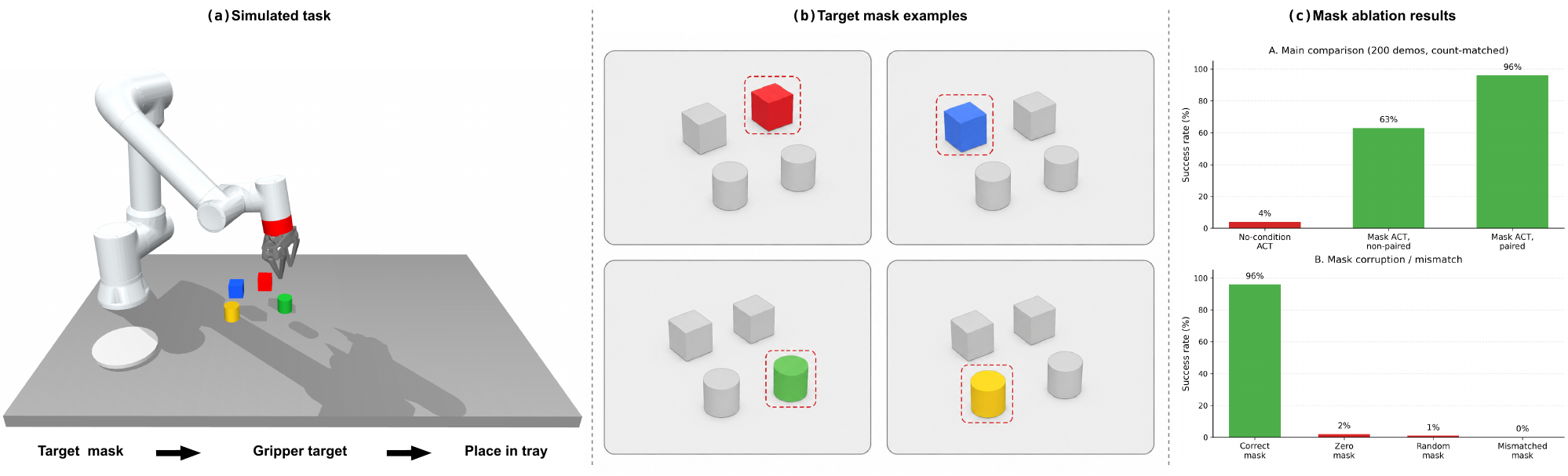}
\caption{Target-conditioned manipulation evaluation. With correct masks, paired-condition mask-conditioned ACT reaches 96\% target success. Zero, random, and mismatched masks reduce original-target success to 0--2\%, indicating that the learned behavior depends on the provided target condition.}
\label{fig:manipulation}
\end{figure}

Before evaluating the autonomous excavation performance, a diagnostic manipulation task (shown in Fig.~\ref{fig:manipulation}) is introduced to isolate and verify whether the learned policy correctly leverages the target condition.
The task is target-conditioned grasping in MuJoCo with a Fairino FR5 robot and an OmniPicker adaptive gripper.
Each scene contains four blocks, and the target is specified by an image-aligned binary mask.
The policy receives RGB+mask observations and proprioception and predicts action chunks of length 100 at 50 Hz.

All main comparisons use 200 demonstrations and 50 closed-loop rollout evaluations per setting.
The paired-condition dataset contains 50 scenes with four target episodes per scene.
Episodes are randomly split into 80\% training and 20\% validation for model selection.
The strict success criterion requires grasping the block specified by the target mask and placing it into a side tray.

\begin{table}[t]
\caption{Target-conditioned manipulation results. Correct-mask rows compare no-condition, non-paired mask-conditioned, and paired-condition mask-conditioned policies. Mask-corruption rows evaluate the paired model and report original-target success.}
\label{tab:manipulation}
\centering
\small
\resizebox{\linewidth}{!}{%
\begin{tabular}{llcrr}
\toprule
Setting & Test mask & Demos & Success & Avg return \\
\midrule
No-condition ACT & None & 200 & 4\% & 32.06 \\
Mask-conditioned ACT, non-paired & Correct & 200 & 63\% & 448 \\
Mask-conditioned ACT, paired & Correct & 200 & 96\% & 658 \\
Mask-conditioned ACT, paired & Zero & 200 & 2\% & 0 \\
Mask-conditioned ACT, paired & Random & 200 & 1\% & 12.5 \\
Mask-conditioned ACT, paired & Mismatched & 200 & 0\% & 0 \\
\bottomrule
\end{tabular}
}
\end{table}

Table~\ref{tab:manipulation} shows that no-condition ACT reaches only 4\% target success despite using the same 200 paired demonstrations.
Providing a mask without paired target contrasts improves success to 63\%, while paired-condition mask-conditioned ACT reaches 96\%.
Mask corruptions reduce original-target success to 0--2\%, which supports the conclusion that the policy complies with the specified visual target rather than executing a memorized default grasp.

\subsection{Excavation Simulation Results}

Table~\ref{tab:sim_results} reports the excavation simulation results.
In single-scoop trials, no-condition ACT can learn a useful default dig-load-dump primitive, while mask-conditioned ACT with paired data achieves 46/50 target success.
The stronger test is sequential pile clearing, where the system must allocate repeated scoops across the pile.
In that setting, paired-condition mask-conditioned ACT removes 76.8\% of the pile, compared with 27.4\% for non-paired mask-conditioned ACT and 15.7\% for no-condition ACT.

\begin{table*}[t]
\caption{Excavation simulation results. Single-scoop trials evaluate one dig-load-dump primitive. Sequential pile-clearing trials evaluate target-conditioned allocation of repeated scoops over a pile. Human-normalized efficiency is computed relative to joystick demonstrations under the matched protocol.}
\label{tab:sim_results}
\centering
\small
\begin{tabular}{llccccc}
\toprule
Protocol & Method & Trials & Target Succ. & Retained Mass & Pile Removed & Human-Norm. Eff. \\
\midrule
Single scoop & No-condition ACT & 50 & n/a & $418 \pm 82$ kg & n/a & $86.4 \pm 9.1\%$ \\
Single scoop & Mask-conditioned ACT, paired & 50 & 46/50 (92.0\%) & $405 \pm 88$ kg & n/a & $83.5 \pm 8.7\%$ \\
Single scoop & Joystick demonstrations & 50 & -- & $485 \pm 65$ kg & n/a & $100\%$ \\
Pile clearing & No-condition ACT & 10 & n/a & -- & $15.7 \pm 4.3\%$ & $18.6 \pm 5.1\%$ \\
Pile clearing & Mask-conditioned ACT, non-paired & 10 & -- & -- & $27.4 \pm 6.2\%$ & $32.5 \pm 7.4\%$ \\
Pile clearing & Mask-conditioned ACT, paired & 10 & -- & -- & $76.8 \pm 7.1\%$ & $91.0 \pm 8.4\%$ \\
Pile clearing & Joystick demonstrations & 10 & -- & -- & $84.4 \pm 5.8\%$ & $100\%$ \\
\bottomrule
\end{tabular}
\end{table*}

These results support a practical intelligent-control conclusion: target conditioning matters most when local excavation actions must be distributed across a larger work area.
A no-condition policy can perform a repeated default scoop, but it lacks an input channel for selecting the next dig region.
Paired-condition mask-conditioned ACT provides the strongest sequential pile-clearing result under the tested simulation protocol.

\section{Discussion}

The manipulation study verifies whether the controller follows the visual command, the single-scoop protocol evaluates the dig-load-dump primitive, and pile clearing tests whether repeated scoops cover a larger work area.
The results show that target conditioning matters most for this final capability: a no-condition policy can retain useful mass in one scoop but cannot select the next dig region.
Paired-condition demonstrations improve sequential behavior by exposing target-dependent action variation during training.

The framework is a simulation-based validation study rather than a claim of full-size construction-site autonomy.
Quantitative material-removal comparisons are reported in simulation under controlled pile states and measurements.
The scaled testbed verifies interface-level closed-loop execution but not calibrated material-removal performance or direct deployment to full-size machinery.
Future work should evaluate more target generators, broader soil and pile distributions, hardware actuation limits, and failure categories such as spill-before-target, hard target collision, incomplete dumping, and repeated default scooping.

\section{Conclusion}

This paper presented a target-conditioned intelligent control framework for autonomous excavation.
The framework connects target masks, joystick demonstrations, action-chunk policy execution, and physics-based deformable-soil simulation.
Paired-condition supervision provides target-action contrasts that help the policy use the commanded region rather than execute a default behavior.
Diagnostic manipulation results verify condition-dependent behavior, and excavation simulation results show that paired-condition mask-conditioned ACT improves sequential pile clearing over no-condition and non-paired baselines.
These findings support target-conditioned action chunking as a practical intelligent-control implementation for simulation-based excavator automation.

\bibliographystyle{ieeetr}
\bibliography{references}

\end{document}